\documentclass[10pt,twocolumn,letterpaper]{article}

\usepackage[pagenumbers]{cvpr}

\definecolor{cvprblue}{rgb}{0.21,0.49,0.74}
\usepackage[pagebackref,breaklinks,colorlinks,citecolor=cvprblue]{hyperref}

\usepackage{adjustbox}
\usepackage{multirow}
\usepackage{amssymb}
\usepackage{bbding}
\usepackage{makecell}

\title{SceneGen: Single-Image 3D Scene Generation in One Feedforward Pass}

\author{Yanxu Meng$^{*}$,\, Haoning Wu$^{*}$,\, Ya Zhang,\, Weidi Xie \\[4pt]
School of Artificial Intelligence, Shanghai Jiao Tong University \\[2pt]
}

\begin{document}

\twocolumn[{%
\renewcommand\twocolumn[1][]{#1}%
\maketitle
\vspace{-18pt}
\begin{center}
   \centering
   \includegraphics[width=.99\textwidth]{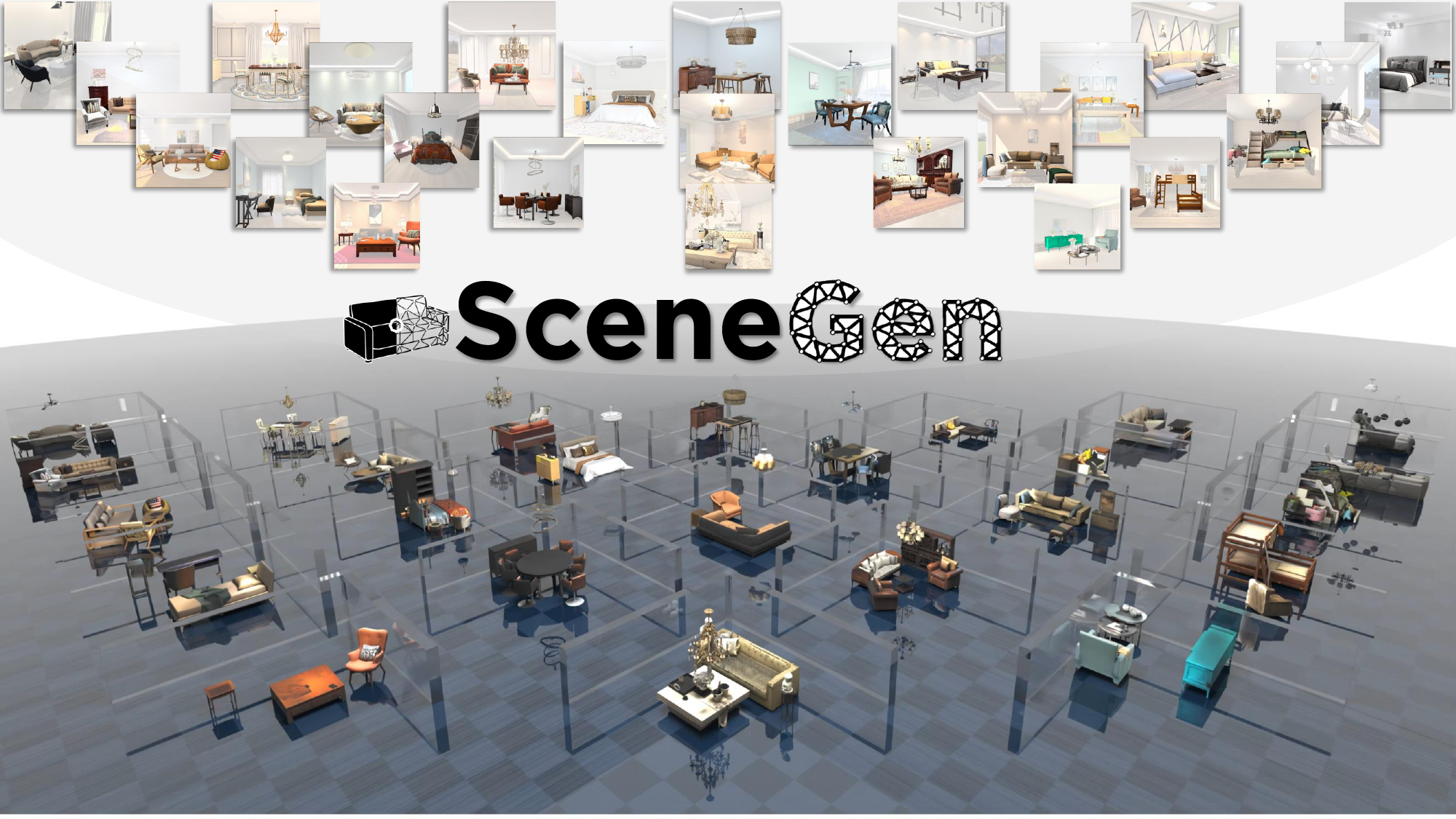}
   \captionof{figure}{
   \textbf{Overview.}
   Our proposed \textbf{SceneGen} framework takes a single scene image and its corresponding object masks as inputs, and efficiently generates multiple 3D assets with coherent geometry, texture, and spatial arrangement in a single feedforward pass.
   }
  \label{fig:teaser}
 \end{center}
}]

\renewcommand{\thefootnote}{}
\footnotetext[1]{\hspace{-0.5cm} *: These authors contribute equally to this work.}

\begin{abstract}
3D content generation has recently attracted significant research interest, driven by its critical applications in VR/AR and embodied AI. 
In this work, we tackle the challenging task of \textbf{synthesizing multiple 3D assets within a single scene image}. 
Concretely, our contributions are fourfold:
(i) we present \textbf{SceneGen}, a novel framework that takes a scene image and corresponding object masks as input, simultaneously producing multiple 3D assets with geometry and texture. 
Notably, SceneGen operates with no need for extra optimization or asset retrieval;
(ii) we introduce a novel \textbf{feature aggregation} module that integrates local and global scene information from visual and geometric encoders within the \textbf{feature extraction} module.
Coupled with a \textbf{position head}, this enables the generation of 3D assets and their relative spatial positions in a single feedforward pass;
(iii) we demonstrate SceneGen's direct extensibility to multi-image input scenarios. 
Despite being trained solely on single-image inputs, our architecture yields improved generation performance when multiple images are provided;
and
(iv) extensive quantitative and qualitative evaluations confirm the efficiency and robustness of our approach.
We believe this paradigm offers a novel solution for high-quality 3D content generation, potentially advancing its practical applications in downstream tasks. 
The code and model will be publicly available at: \url{https://mengmouxu.github.io/SceneGen}.

\end{abstract}

\vspace{-6pt}
{\em ``Everything you can imagine is real."}

{\em \hspace{5.0cm}------ Pablo Picasso}
    
\section{Introduction}
\label{sec:intro}
The growing demand for immersive digital environments in applications such as virtual/augmented reality~(VR/AR) and embodied AI has spurred significant advancements in 3D content generation~\cite{fu20203dfuture, fu20213dfront, deitke2023objaverse, deitke2023objaverse-xl, collins2022abo, khanna2024hssd}. 
While early efforts primarily focus on synthesizing individual 3D assets~\cite{trellis, li2025triposg, zhang2024clay}, recent research attention has shifted to the more challenging task of 3D scene generation. 
Generating realistic 3D scenes~\cite{chu2023buol, gao2024diffcad, gumeli2022roca, yu2025metascenes, zhai2023commonscenes, dong2025hiscene, gao2024graph}, whether conditioned on input text or images, requires synthesizing multiple assets with accurate geometry, texture, and spatial relationships. 
This challenge fundamentally hinges on two key capabilities:
(i) \textbf{3D asset generation} for creating plausible asset geometric topologies from limited textual or visual input;
and 
(ii) \textbf{spatial arrangement} for managing inter-object spatial relationships to ensure physical plausibility, such as support, occlusion, and other interactions among assets.

In general, existing works fall into two paradigms:
(i) retrieval-based methods~\cite{sun2025layoutvlm, feng2023layoutgpt, yang2024holodeck, liu2025agentic} typically employ LLMs for layout planning and retrieve matching 3D assets from existing libraries to assemble scenes. 
Though straightforward, their flexibility is constrained by the coverage of available asset libraries;
(ii) two-stage approaches~\cite{gu2025artiscene, ling2025scenethesis, yao2025cast} first synthesize individual 3D assets and then employ vision-language models~(VLMs) or optimization techniques to refine scene structure and spatial arrangement. 
While more flexible, their reliance on iterative optimization inevitably leads to inefficiency and error accumulation. 
The most relevant works to ours are PartCrafter~\cite{lin2025partcrafter} and MIDI~\cite{huang2025midi}, which generate parts or multiple assets from a single image.
However, they still suffer from limited synthesis fidelity and inaccurate spatial relations among assets.

To tackle the aforementioned challenges, we propose \textbf{SceneGen}, a novel 3D scene generation model designed to simultaneously generate multiple assets, including their geometry, texture, and spatial positions, from a single scene image in a single feedforward pass~(Figure~\ref{fig:teaser}).
Concretely, our framework builds upon an existing single-asset generation model~\cite{trellis} and incorporates three key modules: \textbf{feature extraction}, \textbf{feature aggregation}, and \textbf{output}.

Specifically, the feature extraction module first strategically leverages off-the-shelf visual~\cite{dinov2} and geometric~\cite{wang2025vggt} encoders to extract both asset-level and scene-level representations.
Subsequently, our proposed feature aggregation module, composed of local and global attention blocks, effectively integrates these visual and geometric features while facilitating inter-asset interactions during generation to ensure geometrically plausible topologies.
Finally, leveraging this comprehensive scene context, the output module can directly decode the generated latent features into the assets' relative position, geometry, and texture via a position head and a pre-trained structure decoder.

Moreover, despite being trained exclusively on single-image samples, SceneGen exhibits remarkable generalization to multi-image input scenarios, yielding even better generation quality, which primarily stems from our dedicated architectural design. 
To ensure a comprehensive and reliable evaluation of SceneGen, we systematically adopt multiple metrics focusing on both geometric and visual quality. 
Both quantitative and qualitative results demonstrate that our proposed SceneGen significantly outperforms previous methods in terms of generation quality and efficiency, which can generate a textured scene with four assets in approximately 2 minutes on a single A100 GPU.

The rest of this paper is organized as follows:
Sec.~\ref{sec:related_work} provides a comprehensive review and discussion of related literature.
Sec.~\ref{sec:method} elaborates on our proposed SceneGen framework.
Sec.~\ref{sec:experiments} presents extensive quantitative and qualitative evaluations.
Finally, Sec.~\ref{sec:conclusion} concludes with key insights and contributions.
To our knowledge, SceneGen is the first 3D scene generation model capable of \textbf{simultaneously synthesizing geometry, texture, and relative positions of multiple 3D assets in a single feedforward pass, without requiring per-scene optimization}.
We believe this work will inspire future advances in high-quality, efficient 3D content generation and facilitate diverse downstream applications.

\section{Related Work}
\label{sec:related_work}
\noindent \textbf{3D visual perception.}
Extensive research has advanced 3D visual perception, where traditional methods like SfM~\cite{schonberger2016structure, wang2024vggsfm} rely on computationally intensive optimization for 3D reconstruction.
Notably, emerging feedforward methods~\cite{wang2024dust3r, leroy2024mast3r, zhang2025monst3r, wang2025continuous, wang2025vggt, chen2025easi3r, zhang2025flare, jiang2025geo4d, team2025aether} have demonstrated efficient 3D perception, with DUSt3R~\cite{wang2024dust3r} pioneering this trend and VGGT~\cite{wang2025vggt} establishing a minimalist yet powerful paradigm that distills geometric priors from large-scale data without explicit 3D inductive biases or optimizations.

\vspace{2pt}
\noindent \textbf{3D asset synthesis.}
Typically, 3D asset synthesis aims to generate object-centric geometry and texture from text or image inputs.
The recent success of diffusion models~\cite{ho2020ddpm} in 2D generation~\cite{SDM, peebles2023DiT, liu2024intelligent, wu2024megafusion, wu2025mrgen} has inspired the development of learning-based, scalable 3D content~\cite{fu20203dfuture, fu20213dfront, deitke2023objaverse, deitke2023objaverse-xl, collins2022abo, khanna2024hssd} generation, which produce 3D asset in various representations, including explicit forms such as point clouds~\cite{luo2021pointcloud}, voxels~\cite{muller2023diffrf, hui2022voxel}, and SDFs~\cite{cheng2023sdf, li2023sdf}, as well as implicit ones like 3D Gaussians~\cite{he2024gvgen, zhang2024gaussiancube} and NeRFs~\cite{anciukevivcius2023nerf, li2023instant3d, xu2023dmv3d}.
Subsequent advances leverage VAEs~\cite{kingma2013VAE} for compressing 3D geometry or textures~\cite{trellis, li2025triposg, zhang2024clay} and adopt hybrid mesh-texture pipelines~\cite{wu2024direct3d, zhang20233dshape2vecset, huang2024mv, hunyuan3d2025hunyuan3d21imageshighfidelity}, with TRELLIS~\cite{trellis} demonstrating scalable, high-fidelity generation via structured latents. 
Nevertheless, these methods remain restricted to single-asset synthesis and fundamentally lack the capability to model complex multi-asset scenes.

\begin{figure}[t]
  \centering
  \includegraphics[width=.47\textwidth]{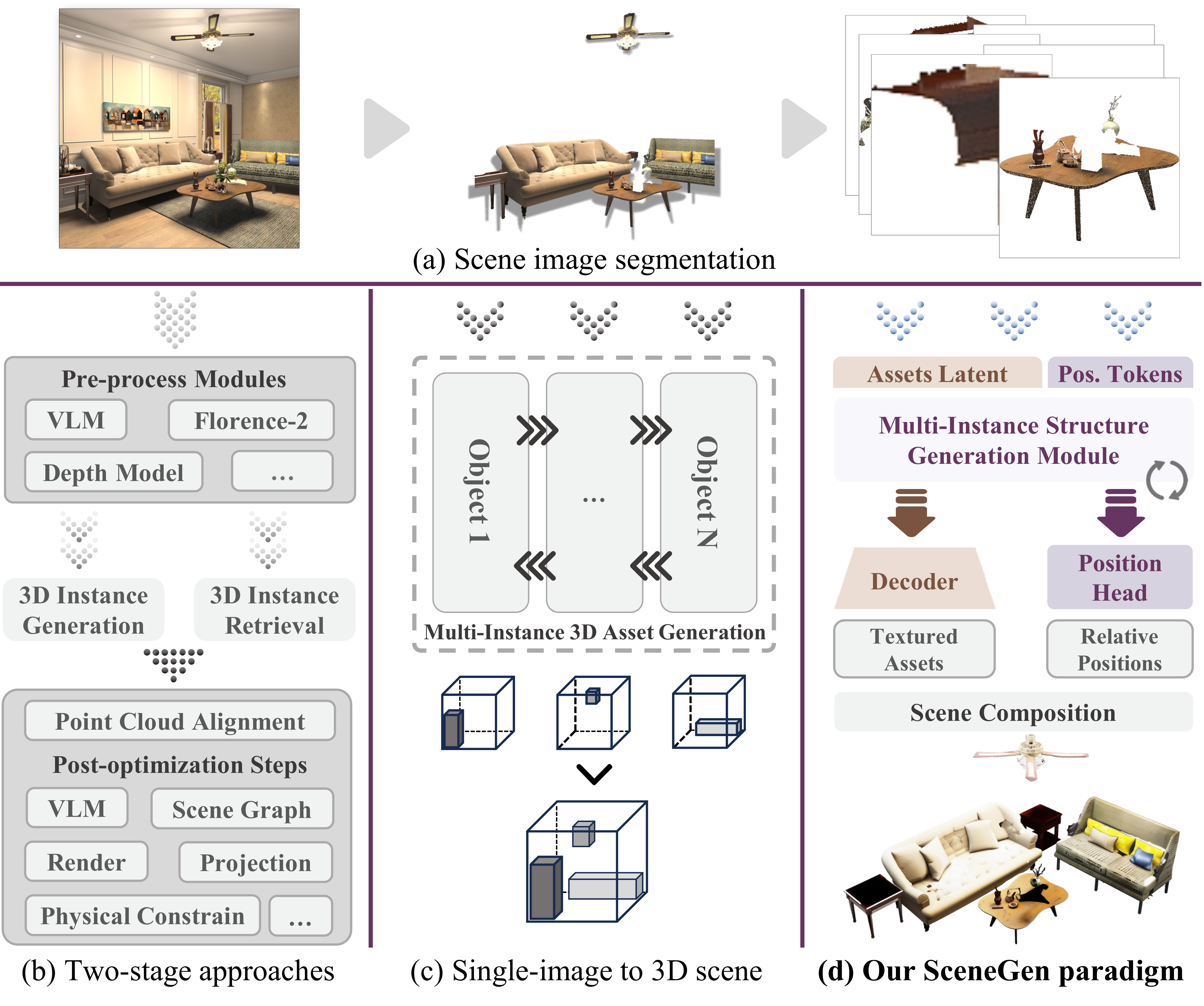} \\
  \vspace{-2pt}
  \caption{
    \textbf{3D Scene Generation.}
    (a) Existing methods typically require segmenting target objects from the scene image;
    (b) Two-stage methods like CAST~\cite{yao2025cast} sequentially retrieve or generate individual assets, then assemble them via post-processing;
    (c) Methods such as MIDI~\cite{huang2025midi} directly generate multiple assets from a single image, but suffer from blurry details and unreasonable spatial layouts;
    (d) In contrast, our SceneGen jointly synthesizes the geometry, texture, and spatial positions of multiple assets in a single feedforward pass, producing plausible 3D scenes. 
  }
  \vspace{-8pt}
  \label{fig:pipeline}
\end{figure}

\vspace{2pt}
\noindent \textbf{3D scene generation.}
Beyond single-asset synthesis, 3D scene generation is more challenging yet valuable, aiming to produce multiple coordinated, physically plausible assets within a scene.
Prior text-based approaches primarily leverage LLMs for layout planning~\cite{sun2025layoutvlm, feng2023layoutgpt, yang2024holodeck, liu2025agentic} and retrieve suitable assets from existing libraries.
Subsequent image-based methods employ segmentation~\cite{chu2023buol, gao2024diffcad, gumeli2022roca, yu2025metascenes, dogaru2025gen3dsr}, scene graphs~\cite{zhai2023commonscenes, dong2025hiscene, gao2024graph} and depth/point cloud alignment~\cite{tang2025towards, yao2025cast, dogaru2025gen3dsr} to assist in multi-asset generation and arrangement.
As depicted in Figure~\ref{fig:pipeline}~(b), recent optimization-based methods~\cite{gu2025artiscene, ling2025scenethesis, yao2025cast} adopt VLMs for post-processing, refining scene structures via image- or text-guided adjustments, but inevitably suffer from inefficiency.
Other works~(Figure~\ref{fig:pipeline}~(c)), such as MIDI~\cite{huang2025midi} and PartCrafter~\cite{lin2025partcrafter}, explore scene generation conditioned on a single image, but inherently sacrifice reconstruction fidelity due to their reliance on canonical-space representations.
To overcome these limitations, our proposed \textbf{SceneGen} uniquely integrates asset-level and scene-level features, enabling robust and efficient 3D scene generation.

\section{Method}
\label{sec:method}
In this work, we present \textbf{SceneGen}, designed to jointly perform 3D asset generation within scenes and predict relative spatial positions among assets. 
Here, we first formally describe our problem formulation in Sec.~\ref{subsec:problem_formulation}; followed by elaboration on our model architecture and training method in Sec.~\ref{subsec:scenegen} and Sec.~\ref{subsec:training}, respectively; finally, we extend SceneGen to multi-view input scenarios in Sec.~\ref{subsec:multi-view_scene_generation}.

\subsection{Problem Formulation}
\label{subsec:problem_formulation}
Our proposed \textbf{SceneGen} is a single-stage feedforward 3D scene generation model~($\boldsymbol{\mathcal{G}}_{\mathrm{Scene}}$), which takes a scene image~($\boldsymbol{I}_\mathrm{Scene}$) containing $N$ objects and corresponding masks~($\{\boldsymbol{m}_i\}_{i=1}^{N}$) as input, simultaneously generating 3D asset structure and texture representations~($\{\boldsymbol{S}_i\}_{i=1}^{N}$), 
and their relative positions~($\{\boldsymbol{P}_i\}_{i=1}^{N}$), formulated as:
\begin{equation*}
    \{(\boldsymbol{S}_i, \boldsymbol{P}_i)\}_{i=1}^N = \boldsymbol{\mathcal{G}}_{\mathrm{Scene}}(\boldsymbol{I}_\mathrm{Scene}, \{\boldsymbol{m}_i\}_{i=1}^N)
\end{equation*}
Here, the position of each asset relative to a pre-selected query asset, is denoted as $\boldsymbol{P}_i = [\boldsymbol{t}_i, \boldsymbol{q}_i, \boldsymbol{s}_i] \in \mathbb{R}^8$, comprising $\boldsymbol{t}_i \in \mathbb{R}^3$~(translation), $\boldsymbol{q}_i \in \mathbb{R}^4$~(rotation quaternion), and $\boldsymbol{s}_i \in \mathbb{R}^1$~(scale factor).
By default, we select the asset with $i=1$ as the query asset, with its parameters fixed as: $\boldsymbol{t}_{\mathrm{query}} = [0,0,0]$, $\boldsymbol{q}_{\mathrm{query}} = [1,0,0,0]$, $\boldsymbol{s}_{\mathrm{query}} = 1$.

\begin{figure*}[t]
  \centering
  \includegraphics[width=\textwidth]{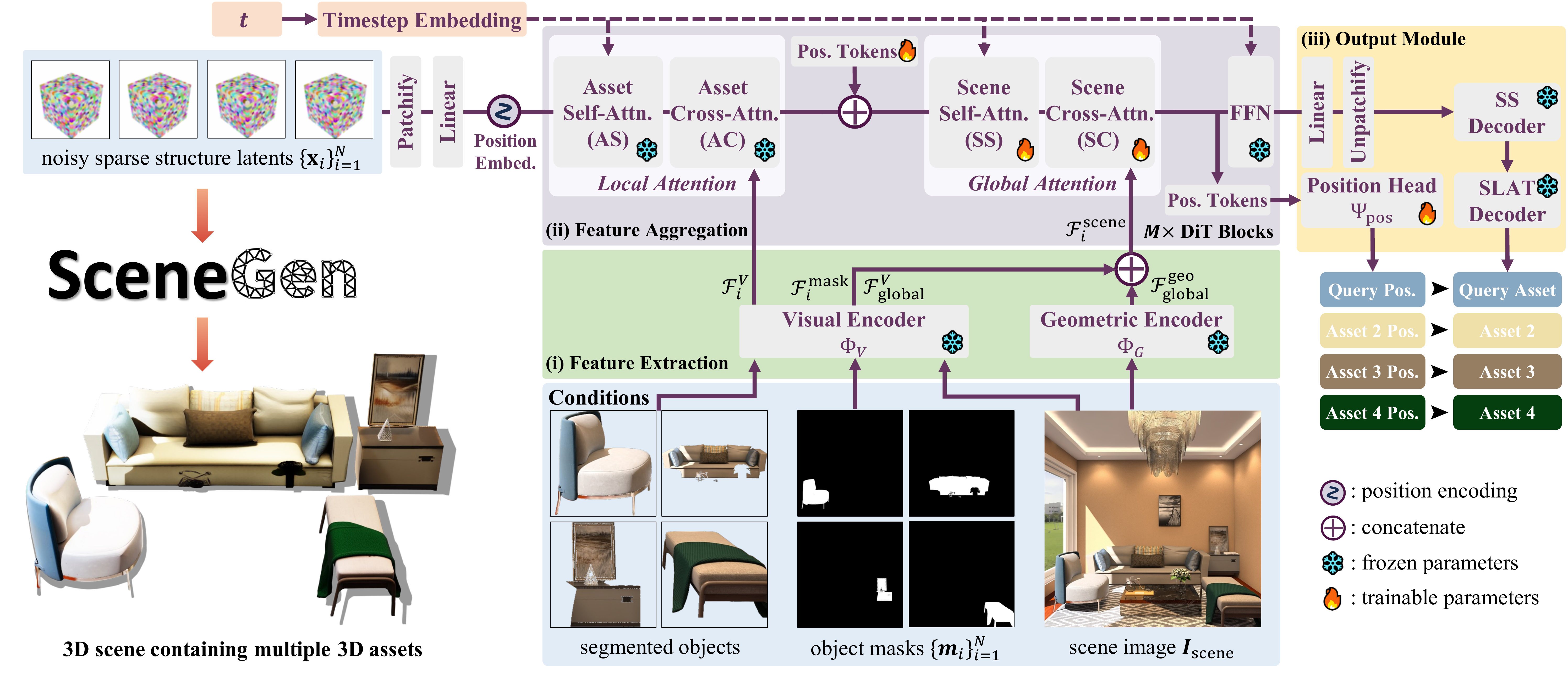} \\
  \vspace{-2pt}
  \caption{
    \textbf{Architecture Overview.}
    \textbf{SceneGen} takes a single scene image with multiple objects and corresponding segmentation masks as input.
    A pre-trained {\em local attention} block first refines the texture of each asset.
    Then, our introduced {\em global attention} block integrates asset-level and scene-level features extracted by dedicated visual and geometric encoders.
    Finally, two off-the-shelf structure decoders and our {\em position head} decode these latent features into multiple 3D assets with geometry, texture, and relative spatial positions.
  }
    \vspace{-8pt}
 \label{fig:arch}
\end{figure*}

\subsection{SceneGen}
\label{subsec:scenegen}
SceneGen framework~($\boldsymbol{\mathcal{G}}_{\mathrm{Scene}}$) comprises three key stages: 
(i) \textbf{feature extraction}, employing a scene visual encoder~($\mathrm{\Phi}_V$) and a scene geometric encoder~($\mathrm{\Phi}_G$) to extract visual and structural features within the scene, implemented using pre-trained DINOv2~\cite{dinov2} and VGGT~\cite{wang2025vggt}, respectively;
(ii) \textbf{feature aggregation}, comprising $M$ DiT~\cite{peebles2023DiT} blocks, each integrating a {\em local attention} block, a {\em global attention} block, and a feedforward network;
and 
(iii) \textbf{output} module, which introduces a position head~($\Psi_{\mathrm{pos}}$) for predicting the spatial locations of assets and adopts off-the-shelf sparse structure~(SS) and structured latents~(SLAT) decoders~\cite{trellis} for decoding scene geometry structures.
By integrating these complementary modules, our SceneGen effectively captures both local asset-level and global scene-level features, enabling it to simultaneously generate multiple 3D assets and predict their relative positions. 

\vspace{2pt}
\noindent \textbf{Feature extraction.}
SceneGen starts with extracting both local and global features from a given scene image~($\boldsymbol{I}_{\mathrm{scene}}$) with the visual encoder~($\Phi_{V}$) and geometric encoder~($\Phi_{G}$). Specifically, for each object with its corresponding segmentation mask~($\boldsymbol{m}_i$), we obtain four complementary feature representations:
(i) the object's individual visual features~($\mathcal{F}_i^{V}$);
(ii) the visual features of its mask~($\mathcal{F}_i^{\mathrm{mask}}$);
(iii) scene global visual features~($\mathcal{F}_{\mathrm{global}}^{V}$);
and
(iv) the global geometric features~($\mathcal{F}_{\mathrm{global}}^\mathrm{{geo}}$), formulated as:
\begin{align*}
    & \mathcal{F}_{i}^{V} = \Phi_{V}(\boldsymbol{I}_{\mathrm{scene}}\otimes \boldsymbol{m}_i),  \quad \mathcal{F}_i^{\mathrm{mask}} = \Phi_{V}(\boldsymbol{m}_i), \\
    & \mathcal{F}_{\mathrm{global}}^{V} = \Phi_{V}(\boldsymbol{I}_{\mathrm{scene}}), \quad\, \mathcal{F}_{\mathrm{global}}^\mathrm{{geo}} = \Phi_{G}(\boldsymbol{I}_{\mathrm{scene}})
\end{align*}
Here, $\otimes$ denotes pixel-wise multiplication.
These features are then concatenated along the sequence length dimension into a unified scene context~($\mathcal{F}_i^{\mathrm{scene}}$), formulated as:
\begin{align*}
    \mathcal{F}_i^{\mathrm{scene}} = [\mathcal{F}_i^{V}; \mathcal{F}_i^{\mathrm{mask}}; \mathcal{F}_{\mathrm{global}}^{V}; \mathcal{F}_{\mathrm{global}}^\mathrm{{geo}}]
\end{align*}
%

\noindent \textbf{Feature aggregation.}
To integrate the extracted scene context features~($\mathcal{F}_i^{\mathrm{scene}}$), 
SceneGen employs a feature aggregation module, 
that enables the simultaneous generation of multiple 3D assets.
This module comprises a {\em local attention} block that refines details of individual assets, a {\em global attention} block that incorporates scene context to facilitate inter-asset interactions, and a feedforward network.
Concretely, the local attention blocks and feedforward networks are initialised from pre-trained weights from TRELLIS~\cite{trellis}, a flow-matching~\cite{lipman2023flowmatching} model designed to synthesize 3D content from noisy sparse structure latents.
For clarity and conciseness, given the sparse structure latents~($\{\mathbf{x}_i\}_{i=1}^N$, where each ${\mathbf{x}}_i \in \mathbb{R}^{T \times C}$) of $N$ objects in a scene, we denote the standard attention mechanism as $\mathrm{Attention}(\mathbf{Q}, \mathbf{K}, \mathbf{V})$, and elaborate on a single DiT block as follows.

The {\em local attention} block aims to enhance details of individual objects through asset-level self-attention~(AS) and cross-attention~(AC).
To be specific, it focuses on fusing the latent features of each object~($\mathbf{x}_i$) with their corresponding visual features~($\mathcal{F}_{i}^{V}$) to yield refined representations of each object~($\mathbf{x}_i^{\mathrm{AC}}$), which can be formulated as:
\begin{align*}
    \mathbf{x}_i^{\mathrm{AS}} & = \mathrm{Attention}(\mathbf{x}_i, \mathbf{x}_i, \mathbf{x}_i) \\
    \mathbf{x}_i^{\mathrm{AC}} & = \mathrm{Attention}(\mathbf{x}_i^{\mathrm{AS}}, \mathcal{F}_{i}^{V}, \mathcal{F}_{i}^{V})
\end{align*}

To establish inter-dependencies among 3D assets, we propose a {\em global attention} block, comprising scene-level self-attention~(SS) and cross-attention~(SC), which capture inter-object relationships and integrate scene geometry, respectively.
Consequently, this design ensures physically plausible spatial arrangements of generated assets.

Similar to~\cite{wang2025vggt}, we initialize one learnable position token~($\boldsymbol{p}_i$) and four register tokens~($\boldsymbol{r}_i$)~\cite{darcet2023register} for refined features of each object~($\mathbf{x}_i^{\mathrm{AC}}$), denoted as:
$\hat{\mathbf{x}}_i = [\boldsymbol{p}_i; \boldsymbol{r}_i; \mathbf{x}_i^{\mathrm{AC}}]$, 
where $[\cdot; \cdot]$ refers to concatenation along the token length dimension.
Notably, we assign a unique position token~($\boldsymbol{p}_\mathrm{query}$) and register tokens~($\boldsymbol{r}_\mathrm{query}$) to the query asset, while adopting shared position token~($\boldsymbol{p}_i$) and register tokens~($\boldsymbol{r}_i$) for other assets.
For each asset feature~($\hat{\mathbf{x}}_i \in \mathbb{R}^{T \times C}$), we concatenate them along the token sequence dimension to form a unified scene representation~($\mathbf{X} \in \mathbb{R}^{(N \cdot T) \times C}$), which is processed by our scene-level self-attention layer, resulting in updated tokens of each asset~($\{\mathbf{x}_i^{\mathrm{SS}}\}_{i=1}^N$), formulated as:
\begin{align*}
    \{\mathbf{x}_i^{\mathrm{SS}}\}_{i=1}^N = \mathrm{Attention}(\mathbf{X}, \mathbf{X}, \mathbf{X})
\end{align*}
Through this process, intra-asset and inter-asset information aggregation establishes essential shape and position awareness for coherent multi-asset generation.
We then employ scene-level cross-attention to integrate multiple pre-extracted scene-aware features, thus incorporating 3D geometric context.
The features of each asset are updated into geometry-aware representations~($\{\mathbf{x}_i^{\mathrm{SC}}\}_{i=1}^N$), denoted as:
\begin{align*}
     \mathbf{x}_i^{\mathrm{SC}} = \mathrm{Attention}(\mathbf{x}_i^{\mathrm{SS}}, \mathcal{F}_i^{\mathrm{scene}}, \mathcal{F}_i^{\mathrm{scene}})
\end{align*}
This preserves object-specific details while integrating global geometric constraints, which effectively addresses occlusion challenges and enables geometric refinement.

\vspace{2pt}
\noindent \textbf{Output module.}
After passing through $M$ DiT blocks, we obtain the updated position tokens~($\{\hat{\boldsymbol{p}}\}_{i=1}^{N}$) and latent features~($\{\tilde{\mathbf{x}}\}_{i=1}^{N}$) of each generated asset, which are subsequently decoded into their relative spatial positions and detailed 3D representations~(structure and texture), respectively.
For relative positions, we concatenate the position tokens of all non-query assets, which are then decoded into corresponding 8D position vectors~($\{\hat{\boldsymbol{P}}_i\}_{i=2}^{N}$) by our proposed position head~($\Psi_{\mathrm{pos}}$), comprising four self-attention layers and a linear layer, denoted as:
\begin{align*}
    \{\hat{\boldsymbol{P}}_i\}_{i=2}^{N} = \{[\hat{\boldsymbol{t}}_i, \hat{\boldsymbol{q}}_i, \hat{\boldsymbol{s}}_i]\}_{i=2}^{N} = \Psi_{\mathrm{pos}}(\{\hat{\boldsymbol{p}}_i\}_{i=2}^{N})
\end{align*}
Here, each vector~($\hat{\boldsymbol{P}}_i$) represents an asset's spatial position~(translation, rotation, and scale) relative to the pre-selected query asset~($i=1$).
Furthermore, the latent features can be directly decoded into the geometry and texture of each asset~($\{\hat{\boldsymbol{S}}\}_{i=1}^{N}$) using off-the-shelf sparse structure generator~($\boldsymbol{\mathcal{G}}_{\mathrm{S}}$) and structured latents generator~($\boldsymbol{\mathcal{G}}_{\mathrm{L}}$) from TRELLIS~\cite{trellis}, formulated as:
\begin{align*}
   \{\hat{\boldsymbol{S}}\}_{i=1}^{N} = \boldsymbol{\mathcal{G}}_{\mathrm{L}}(\boldsymbol{\mathcal{G}}_{\mathrm{S}}(\{\tilde{\mathbf{x}}\}_{i=1}^{N}))
\end{align*}

\subsection{Training}
\label{subsec:training}
During training, only the global attention blocks, learnable position tokens, and position head are optimized, with all other parameters frozen to facilitate efficient training, as depicted in Figure~\ref{fig:arch}.
The technical details regarding training data and loss function designs are presented below.

\vspace{2pt}
\noindent \textbf{Training data.}
Our SceneGen model is trained on the 3D-FUTURE~\cite{fu20203dfuture} dataset, containing photorealistic scene renderings with instance masks and asset annotations.
This dataset comprises 12K training scenes and 4.8K test scenes, each featuring a scene image with one or multiple objects.
To better capture inter-object spatial relationships, we augment the training set by iteratively designating each asset as the query asset while randomly permuting the remaining assets, which expands the effective training samples to 30K.

\vspace{2pt}
\noindent \textbf{Training objectives.}
Our SceneGen model is trained end-to-end using a composite loss function~($\mathcal{L}$) comprising three key components:
(i) the average conditional flow matching~\cite{lipman2023flowmatching} loss~($\mathcal{L}_{\mathrm{cfm}}$), applied to each generated asset for supervising asset generation;
(ii) the position loss~($\mathcal{L}_{\mathrm{pos}}$) for maintaining accurate relative spatial arrangements among assets;
and 
(iii) the voxel-space collision loss~($\mathcal{L}_{\mathrm{coll}}$) for enforcing physically plausible object placement.
The overall objective function~($\mathcal{L}$) combines these components with a weighting factor~($\lambda$), which can be formulated as:
\begin{align*}
    \mathcal{L} = \mathcal{L}_{\mathrm{cfm}} + \lambda(\mathcal{L}_{\mathrm{pos}} + \mathcal{L}_{\mathrm{coll}})
\end{align*}
Concretely, the flow matching loss establishes straight probability paths between distributions via linear interpolation:
$\mathbf{x}_i(t) = (1-t)\mathbf{x}_i^0 + t\boldsymbol{\epsilon}$, where $\boldsymbol{\epsilon}\sim \mathcal{N}(0, \mathbf{I})$, $t \in [0,1]$, and $\mathbf{x}_i^0$ denotes the noise-free sparse structure latents for each of the $N$ assets.
The conditional flow matching objective~($\mathcal{L}_{\mathrm{cfm}}$) learns a parameterized function $\boldsymbol{v}_\theta$ to approximate the velocity field~($\boldsymbol{v}(\mathbf{x}_i(t),t) = \nabla_t\mathbf{x}_i(t)$), represented as:
\begin{align*}
    \mathcal{L}_{\mathrm{cfm}}(\theta) = \frac{1}{N} \sum_{i=1}^{N} \mathbb{E}_{t, \boldsymbol{\epsilon}} \|\boldsymbol{v}_\theta(\mathbf{x}_i(t), t)-(\boldsymbol{\epsilon}-\mathbf{x}_i^0)\|^2_2
\end{align*}
The position loss~($\mathcal{L}_{\mathrm{pos}}$) adopts a $\mu$-weighted Huber loss~($\Vert\cdot \Vert_{\delta_P}$) between the predicted positions~($\hat{\boldsymbol{P}_i} = [\hat{\boldsymbol{t}}_i, \hat{\boldsymbol{q}}_i, \hat{\boldsymbol{s}}_i]$) for all non-query assets~($i \in [2,\dots, N]$) and their ground truth~($\boldsymbol{P}_i = [\boldsymbol{t}_i, \boldsymbol{q}_i, \boldsymbol{s}_i]$), denoted as:
\begin{align*}
    \mathcal{L}_{\mathrm{pos}} = \sum^{N}_{i=2}(& \mu_t\Vert (\hat{\boldsymbol{t}}_i - \boldsymbol{t}_i)/d_{\mathrm{scene}} \Vert_{\delta_P} \\ 
    + & \mu_q\Vert \hat{\boldsymbol{q}}_i - \boldsymbol{q}_i \Vert_{\delta_P} + \mu_s\Vert \hat{\boldsymbol{s}}_i - \boldsymbol{s}_i \Vert_{\delta_P})
\end{align*}
Here, the translation error component is normalized by the scene scale~($d_{\mathrm{scene}}$) of each sample to mitigate numerical instability caused by varying query asset selections.
This stabilizes translation loss during training while improving generalization across distinct query asset configurations.

The collision loss~($\mathcal{L}_{\mathrm{coll}}$) quantifies surface collision in a $64 \times 64 \times 64$ voxel grid~($\boldsymbol{V}$).
Specifically, the predicted sparse structure latents~($\tilde{\mathbf{x}}_i$) are decoded into point clouds~($\{\boldsymbol{p}_i\}_{i=1}^{L}$) via a pre-trained sparse structure decoder from TRELLIS~\cite{trellis}, then transformed using predicted pose parameters~($\hat{\boldsymbol{P}}_i$) and voxelized into $\boldsymbol{V}$.
The collision loss is defined as the ratio of overlapping surface voxels to all surface voxels, using the Huber loss~($\|\cdot\|_{\delta_C}$), denoted as:
\begin{align*}
    \mathcal{L}_{\mathrm{coll}} = \Vert \mathrm{IoU}_{\mathrm{scene}}\Vert_{\delta_C} = \Vert \frac{\sum_i \mathbb{I}[\boldsymbol{V}_i > 1]}{\sum_i \mathbb{I}[\boldsymbol{V}_i > 0]} \Vert_{\delta_C}
\end{align*}
Ideally, $\mathrm{IoU}_{\mathrm{scene}} = 0$ indicates there are no asset collisions.

\subsection{Extension to Multi-view Inputs}
\label{subsec:multi-view_scene_generation}
Despite being trained exclusively on single-image samples, our model exhibits inherent multi-view compatibility, enabled by its flexible feature extraction and conditioning strategy.
Given a scene with $K$ input views~($\{\boldsymbol{I}_\mathrm{scene}^k\}_{k=1}^{K}$), the visual features~($\mathcal{F}_{V}^k$) for each view are extracted independently via the visual encoder~($\Phi_{V}$), while the geometric features are derived from a unified scene representations encoded by aggregating information across all views using the geometric encoder~($\Phi_{G}$), denoted as:
\begin{align*}
    \mathcal{F}_{\mathrm{geo}}^k = \Phi_G(\{\boldsymbol{I}_\mathrm{scene}^j\}^K_{j=1})[k]   
\end{align*}
The final asset positions are determined by averaging the predictions across all views. 
Experimental results~(detailed in Sec.~\ref{subsec:qualitative_results}) indicate that this multi-view inference scheme improves generation quality by leveraging better geometric understanding, despite the model having never been explicitly fine-tuned on such multi-view inputs.

\begin{table*}[t]
    \centering
    \setlength{\tabcolsep}{2pt} 
    \renewcommand{\arraystretch}{1.05} 
    \resizebox{\linewidth}{!}{
        \begin{tabular}{lc|ccccc|c|cccccc|c}
        \toprule[1.5pt]
        \multirow{2}{*}{\centering \makecell{\textbf{Method}}} & 
        \multirow{2}{*}{\centering \makecell{\textbf{Instance} \\ \textbf{Specific}}} & 
        \multicolumn{5}{c|}{\textbf{Geometric Metrics}} & 
        \multirow{2}{*}{\centering \makecell{\textbf{Image} \\ \textbf{Category}}} & 
        \multicolumn{6}{c|}{\textbf{Visual Metrics}} & \multirow{2}{*}{\centering \makecell{\textbf{Inference} \\ \textbf{Time~(s)}}}
        \\
        \cline{3-7} \cline{9-14}
         & & \textbf{CD-S}$\downarrow$ & \textbf{CD-O}$\downarrow$ & \textbf{F-Score-S}$\uparrow$ & \textbf{F-Score-O}$\uparrow$ & \textbf{IoU-B}$\uparrow$ & & \textbf{PSNR}$\uparrow$ & \textbf{SSIM}$\uparrow$ & \textbf{LPIPS}$\downarrow$ & \textbf{FID}$\downarrow$ & \textbf{CLIP-S}$\uparrow$ & \textbf{DINO-S}$\uparrow$ & \\
        \midrule
        \multirow{2}{*}{\centering \makecell{PartCrafter~\cite{lin2025partcrafter}}} & \multirow{2}{*}{\centering \makecell{\XSolidBrush}} & \multirow{2}{*}{\centering \makecell{0.2027}} & \multirow{2}{*}{\centering \makecell{---}} & \multirow{2}{*}{\centering \makecell{40.43}} & \multirow{2}{*}{\centering \makecell{---}} & \multirow{2}{*}{\centering \makecell{---}} & \multirow{2}{*}{\centering \makecell{Scene \\ GT-Render}} & --- & --- & --- & --- & --- & --- & \multirow{2}{*}{\centering \makecell{\textbf{7.2}}} \\ 
        & & & & & & & & --- & --- & --- & --- & --- & --- \\ 
    
        \multirow{2}{*}{\centering \makecell{DepR~\cite{zhao2025deprdepthguidedsingleview}}} & \multirow{2}{*}{\centering \makecell{\Checkmark}} & \multirow{2}{*}{\centering \makecell{0.0518}} & \multirow{2}{*}{\centering \makecell{0.0862}} & \multirow{2}{*}{\centering \makecell{63.02}} & \multirow{2}{*}{\centering \makecell{47.66}} & \multirow{2}{*}{\centering \makecell{0.2989}} & \multirow{2}{*}{\centering \makecell{Scene \\ GT-Render}} & --- & --- & --- & --- & --- & --- & \multirow{2}{*}{\centering \makecell{11.6}} \\
        & & & & & & & & --- & --- & --- & --- & --- & --- \\ 
        
        \multirow{2}{*}{\centering \makecell{Gen3DSR~\cite{dogaru2025gen3dsr}}} & \multirow{2}{*}{\centering \makecell{\Checkmark}} & \multirow{2}{*}{\centering \makecell{0.0521}} & \multirow{2}{*}{\centering \makecell{0.0935}} & \multirow{2}{*}{\centering \makecell{61.26}} & \multirow{2}{*}{\centering \makecell{41.26}} & \multirow{2}{*}{\centering \makecell{0.2978}} & \multirow{2}{*}{\centering \makecell{Scene \\ GT-Render}} & 15.92 & 0.8885 & 0.1730 & 63.95 & 0.8059 & 0.4334  & \multirow{2}{*}{\centering \makecell{179.0}} \\
        & & & & & & & & 15.43 & 0.8899 & 0.1660 & 78.26 & 0.7950 & 0.4416 \\ 
    
        \multirow{2}{*}{\centering \makecell{MIDI$^*$~\cite{huang2025midi}}} & \multirow{2}{*}{\centering \makecell{\Checkmark}} & \multirow{2}{*}{\centering \makecell{0.0501}} & \multirow{2}{*}{\centering \makecell{0.0602}} & \multirow{2}{*}{\centering \makecell{68.74}} & \multirow{2}{*}{\centering \makecell{61.04}} & \multirow{2}{*}{\centering \makecell{0.2493}} & \multirow{2}{*}{\centering \makecell{Scene \\ GT-Render}} & \textbf{16.93} & 0.8814 & 0.1778 & 22.75 & 0.8711 & 0.6892 & \multirow{2}{*}{\centering \makecell{42.5}} \\
        & & & & & & & & 15.45 & 0.8814 & 0.1711 & 28.26 & 0.8706 & 0.7034 \\ 
        \midrule
        \multirow{2}{*}{\centering \makecell{\textbf{SceneGen}}} & \multirow{2}{*}{\centering \makecell{\Checkmark}} & \multirow{2}{*}{\centering \makecell{\textbf{0.0118}}} & \multirow{2}{*}{\centering \makecell{\textbf{0.0138}}} & \multirow{2}{*}{\centering \makecell{\textbf{90.60}}} & \multirow{2}{*}{\centering \makecell{\textbf{89.73}}} & \multirow{2}{*}{\centering \makecell{\textbf{0.5818}}} & \multirow{2}{*}{\centering \makecell{Scene \\ GT-Render}} & 16.76 & \textbf{0.8903} & \textbf{0.1417} & \textbf{19.59} & \textbf{0.9152} & \textbf{0.8322} & \multirow{2}{*}{\centering \makecell{26.0}} \\ 
        & & & & & & & & \textbf{17.59} & \textbf{0.8991} & \textbf{0.1234} & \textbf{12.34} & \textbf{0.9236} & \textbf{0.8702} \\ 
        \bottomrule[1.5pt]
        \end{tabular}
        }
        \vspace{-2pt}
        \caption{
        \textbf{Quantitative Comparisons on the 3D-FUTURE Test Set.} 
        We evaluate the geometric structure using scene-level Chamfer Distance~(CD-S) and F-Score~(F-Score-S), object-level Chamfer Distance~(CD-O) and F-Score~(F-Score-O), and volumetric IoU of object bounding boxes~(IoU-B).
        For visual quality, CLIP-S and DINO-S represent CLIP and DINOv2 image-to-image similarity, respectively.
        We report the time cost for generating a single asset on a single A100 GPU, and $^*$ indicates adopting MV-Adapter~\cite{huang2024mv} for texture rendering.
        }
        \label{tab:quantitative_results}
        \vspace{-8pt}
\end{table*}

\section{Experiments}
\label{sec:experiments}
This section starts with the experimental settings in Sec.~\ref{subsec:experimental_settings}, followed by comprehensive quantitative and qualitative evaluations in Sec.~\ref{subsec:quantitative_results} and Sec.~\ref{subsec:qualitative_results}, respectively.
Finally, we conduct ablation studies in Sec.~\ref{subsec:ablation_studies}.

\subsection{Experimental Settings}
\label{subsec:experimental_settings}
\noindent \textbf{Implementation details.}
All experiments are conducted on $8\times$ NVIDIA A100 GPUs, where we train SceneGen for 240 epochs using the AdamW~\cite{loshchilov2018AdamW} optimizer with a learning rate of $5 \times 10^{-5}$ and a batch size of 8.
The weighting factor $\lambda$ decays dynamically within $[0.2, 1]$ using a decay factor of 0.99, and the thresholds of Huber loss $\delta_P$ and $\delta_C$ are set to 0.02 and 0.05, respectively.
To handle varying numbers of assets across training scenes, each training step dynamically samples scenes containing identical asset counts. 
During inference, we adopt 25 sampling steps with the classifier-free guidance~(CFG) weight set to $w=5.0$.

\vspace{2pt}
\noindent \textbf{Evaluation metrics.}
We assess the generated 3D scenes from both geometric and visual perspectives.
For geometry, we reconstruct point clouds from the synthesized asset surfaces and align them with the ground truth using FilterReg~\cite{gao2019filterreg} for faster and more accurate registration than traditional Iterative Closest Point~(ICP~\cite{ICP}).
We then compute commonly used point cloud metrics, Chamfer Distance~(CD) and F-Score, at both scene and object levels, as well as the volumetric IoU of asset bounding boxes.

For visual quality, we focus on the scene texture rendering.
Specifically, after alignment with the ground truth point cloud, we render the predicted scenes with {\em Blender} from the original input camera viewpoint.
We consider two types of ground truth: 
(i) instance-masked scene images extracted using corresponding object masks, 
and 
(ii) images rendered from ground truth assets at the same viewpoint~(excluding ambient lighting). 
We compare our rendered results with both types of ground truth using PSNR, SSIM, LPIPS~\cite{LPIPS}, FID~\cite{heusel2017FID}, CLIP~\cite{CLIP} similarity, and DINOv2~\cite{dinov2} similarity to assess the texture quality of generated assets.
Regarding efficiency, we report the inference time cost for synthesizing a single 3D asset on a single A100 GPU.
More details will be included in Sec.~\ref{subsec:evaluation_protocols} of the \textbf{Appendix}.

\vspace{2pt}
\noindent \textbf{Baselines.}
We compare SceneGen with representative 3D scene generation methods, including PartCrafter~\cite{lin2025partcrafter}, DepR~\cite{zhao2025deprdepthguidedsingleview}, Gen3DSR~\cite{dogaru2025gen3dsr}, and MIDI~\cite{huang2025midi}, using their pre-trained models.
Specifically, we adopt object masks to specify generation targets for all baselines except for PartCrafter, which does not support mask-based control.
Instead, we directly provide PartCrafter with extracted objects and the number of assets as input.
Moreover, as PartCrafter and DepR do not offer code for texture rendering, our evaluation of these methods focuses on geometric quality, while visual quality is compared with Gen3DSR and MIDI~(relying on MV-Adapter~\cite{huang2024mv} for texture synthesis).

\vspace{2pt}
\noindent \textbf{Benchmarks.}
All evaluations are conducted on the 3D-FUTURE~\cite{fu20203dfuture} test set, comprising 4.8K scenes.
Each scene contains a photorealistic rendered image with one or more objects and corresponding segmentation masks as input.

\begin{figure*}[t]
  \centering
  \includegraphics[width=.98\textwidth]{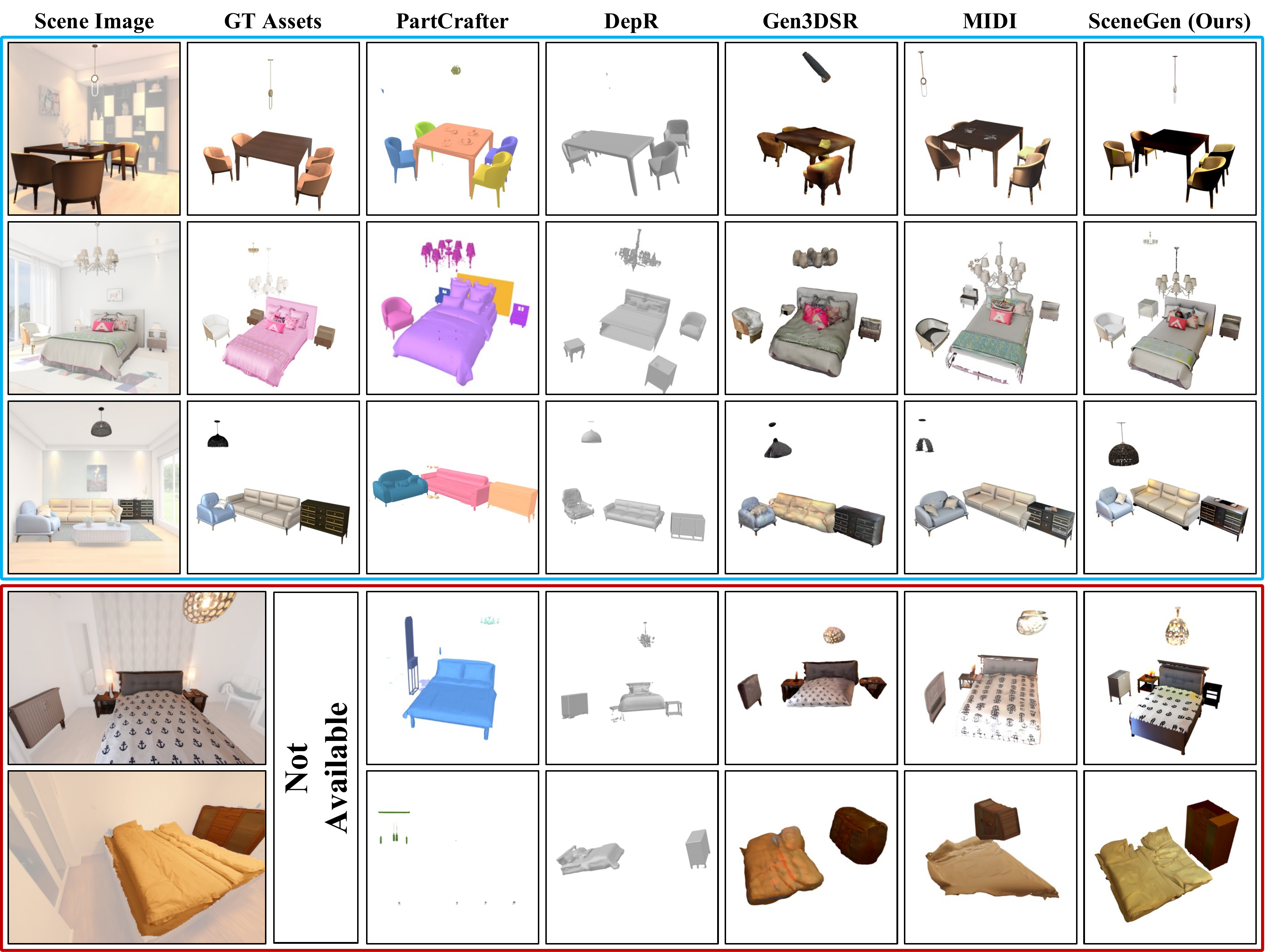} \\
  \vspace{-4pt}
  \caption{
    \textbf{Qualitative Comparisons on the \textcolor[HTML]{00B0F0}{3D FUTURE Test Set} and \textcolor[HTML]{C00000}{ScanNet++}.} 
    Our proposed SceneGen is capable of generating physically plausible 3D scenes featuring \textbf{complete structures}, \textbf{detailed textures}, and \textbf{precise spatial relationships}, demonstrating superior performance over prior methods in terms of both geometric accuracy and visual quality on both the synthetic and real-world datasets.
  }
    \vspace{-12pt}
 \label{fig:qualitative_results}
\end{figure*}

\subsection{Quantitative Results}
\label{subsec:quantitative_results}
As presented in Table~\ref{tab:quantitative_results}, we draw the following key observations:
(i) \textbf{geometric quality}: SceneGen consistently outperforms existing methods across all scene-level and asset-level metrics. 
This stems from its joint integration of local asset features and global scene context during generation.
The interactions among multiple assets facilitate the model in producing physically plausible geometric structures, while the position head further improves the structural realism by explicitly predicting spatial arrangements.
(ii) \textbf{visual quality}: SceneGen can render high-quality textures for generated 3D assets without relying on any external texture generation models.
Moreover, whether using masked scene images or ground-truth renderings as references, our method consistently achieves the best performance across all metrics.
This indicates that our synthesized assets are spatially closer to the ground truth while maintaining superior texture fidelity.
and
(iii) \textbf{efficiency}: While PartCrafter demonstrates a clear advantage in inference speed, it suffers from limited generation quality and controllability.
In contrast, SceneGen achieves both superior quality and a strong balance between quality and efficiency, synthesizing a 3D scene containing four assets with geometry and textures within 2 minutes on a single A100 GPU.

In addition, while the baseline methods, {\em e.g.}, PartCrafter, DepR, and MIDI have been trained on 3D-FRONT~\cite{fu20213dfront}, which may overlap with our test data, our SceneGen still consistently outperforms them across all metrics, further demonstrating its effectiveness and superiority.

\begin{figure}[t]
  \centering
  \includegraphics[width=.47\textwidth]{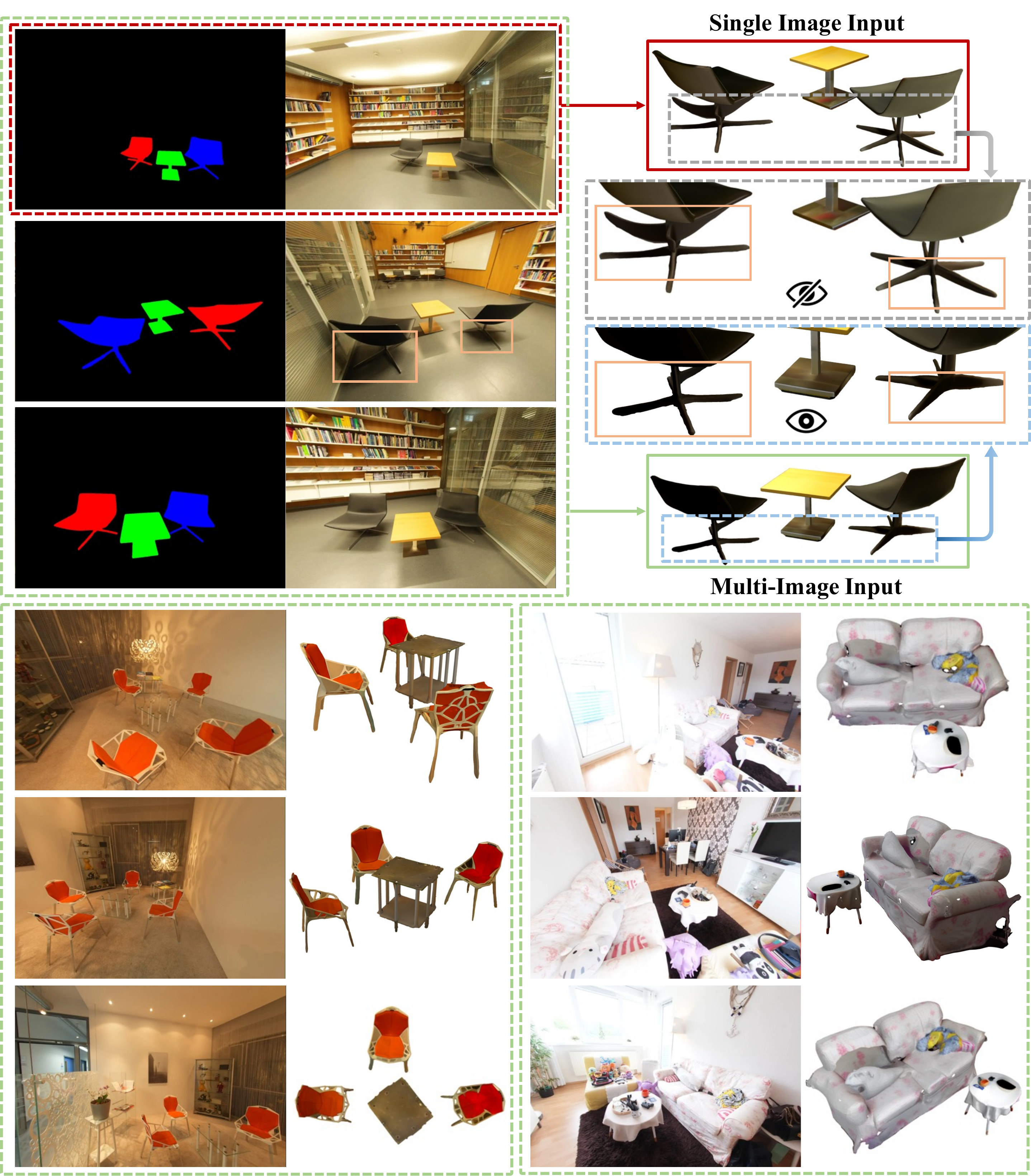} \\
  \vspace{-2pt}
  \caption{
    \textbf{Qualitative Results with Multi-view Inputs.} 
    SceneGen can directly handle multi-view inputs in ScanNet++ and even achieves better generation quality, especially accurate structure.
  }
    \vspace{-8pt}
 \label{fig:extension_to_multi_image}
\end{figure}

\begin{table*}[t]
    \centering
    \setlength{\tabcolsep}{3pt} 
    \renewcommand{\arraystretch}{1.05} 
    \resizebox{\linewidth}{!}{
    \begin{tabular}{cccc|ccccc|c|cccccc}
    \toprule[1.5pt]
    \multirow{2}{*}{\centering \makecell{$\mathcal{F}_{\mathrm{global}}^{\mathrm{geo}}$}} & 
    \multirow{2}{*}{\centering \makecell{$\mathcal{F}_{\mathrm{global}}^{V}$}} & 
    \multirow{2}{*}{\centering \makecell{$\mathcal{F}_i^{\mathrm{mask}}$}} & 
    \multirow{2}{*}{\centering \makecell{$\mathcal{A}_{\mathrm{SS}}$}} & 
    \multicolumn{5}{c|}{\textbf{Geometric Metrics}} & 
    \multirow{2}{*}{\centering \makecell{\textbf{Image} \\ \textbf{Category}}} & 
    \multicolumn{6}{c}{\textbf{Visual Metrics}} 
    \\
    \cline{5-9} \cline{11-16}
     & & & & \textbf{CD-S}$\downarrow$ & \textbf{CD-O}$\downarrow$ & \textbf{F-Score-S}$\uparrow$ & \textbf{F-Score-O}$\uparrow$ & \textbf{IoU-B}$\uparrow$ & & \textbf{PSNR}$\uparrow$ & \textbf{SSIM}$\uparrow$ & \textbf{LPIPS}$\downarrow$ & \textbf{FID}$\downarrow$ & \textbf{CLIP-S}$\uparrow$ & \textbf{DINO-S}$\uparrow$ \\
    \midrule
    \multirow{2}{*}{\centering \makecell{\Checkmark}} & \multirow{2}{*}{\centering \makecell{\Checkmark}} & \multirow{2}{*}{\centering \makecell{\Checkmark}} & \multirow{2}{*}{\centering \makecell{\Checkmark}} & \multirow{2}{*}{\centering \makecell{\textbf{0.0118}}} & \multirow{2}{*}{\centering \makecell{\textbf{0.0138}}} & \multirow{2}{*}{\centering \makecell{\textbf{90.60}}} & \multirow{2}{*}{\centering \makecell{\textbf{89.73}}} & \multirow{2}{*}{\centering \makecell{\textbf{0.5818}}} & \multirow{2}{*}{\centering \makecell{Scene \\ GT-Render}} & \textbf{16.76} & \textbf{0.8903} & \textbf{0.1417} & \textbf{19.59} & \textbf{0.9152} & \textbf{0.8322} \\ 
    & & & & & & & & & & \textbf{17.59} & \textbf{0.8991} & \textbf{0.1234} & \textbf{12.34} & \textbf{0.9236} & \textbf{0.8702} \\ 
    \midrule
    \multirow{2}{*}{\centering \makecell{\XSolidBrush}} & \multirow{2}{*}{\centering \makecell{\Checkmark}} & \multirow{2}{*}{\centering \makecell{\Checkmark}} & \multirow{2}{*}{\centering \makecell{\Checkmark}} & \multirow{2}{*}{\centering \makecell{0.0183}} & \multirow{2}{*}{\centering \makecell{0.0266}} & \multirow{2}{*}{\centering \makecell{83.33}} & \multirow{2}{*}{\centering \makecell{74.71}} & \multirow{2}{*}{\centering \makecell{0.4805}} & \multirow{2}{*}{\centering \makecell{Scene \\ GT-Render}} & 15.89 & 0.8845 & 0.1574 & 20.21 & 0.9049 & 0.8063 \\ 
    & & & & & & & & & & 16.27 & 0.8918 & 0.1421 & 15.36 & 0.9125 & 0.8420 \\ 
    \multirow{2}{*}{\centering \makecell{\XSolidBrush}} & \multirow{2}{*}{\centering \makecell{\XSolidBrush}} & \multirow{2}{*}{\centering \makecell{\Checkmark}} & \multirow{2}{*}{\centering \makecell{\Checkmark}} & \multirow{2}{*}{\centering \makecell{0.0250}} & \multirow{2}{*}{\centering \makecell{0.0286}} & \multirow{2}{*}{\centering \makecell{79.08}} & \multirow{2}{*}{\centering \makecell{73.46}} & \multirow{2}{*}{\centering \makecell{0.4253}} & \multirow{2}{*}{\centering \makecell{Scene \\ GT-Render}} & 15.56 & 0.8806 & 0.1655 & 20.68 & 0.8980 & 0.7850  \\
    & & & & & & & & & & 15.86 & 0.8873 & 0.1511 & 16.62 & 0.9046 & 0.8187 \\ 
    \multirow{2}{*}{\centering \makecell{\XSolidBrush}} & \multirow{2}{*}{\centering \makecell{\XSolidBrush}} & \multirow{2}{*}{\centering \makecell{\XSolidBrush}} & \multirow{2}{*}{\centering \makecell{\Checkmark}} & \multirow{2}{*}{\centering \makecell{0.0310}} & \multirow{2}{*}{\centering \makecell{0.0290}} & \multirow{2}{*}{\centering \makecell{75.20}} & \multirow{2}{*}{\centering \makecell{73.17}} & \multirow{2}{*}{\centering \makecell{0.3825}} & \multirow{2}{*}{\centering \makecell{Scene \\ GT-Render}} & 15.30 & 0.8773 & 0.1730 & 21.12 & 0.8932 & 0.7737  \\
    & & & & & & & & & & 15.55 & 0.8837 & 0.1591 & 17.45 & 0.9000 & 0.8076 \\ 
    \multirow{2}{*}{\centering \makecell{\XSolidBrush}} & \multirow{2}{*}{\centering \makecell{\XSolidBrush}} & \multirow{2}{*}{\centering \makecell{\XSolidBrush}} & \multirow{2}{*}{\centering \makecell{\XSolidBrush}} & \multirow{2}{*}{\centering \makecell{0.0764}} & \multirow{2}{*}{\centering \makecell{0.0352}} & \multirow{2}{*}{\centering \makecell{54.21}} & \multirow{2}{*}{\centering \makecell{70.55}} & \multirow{2}{*}{\centering \makecell{0.1705}} & \multirow{2}{*}{\centering \makecell{Scene \\ GT-Render}} & 13.32 & 0.8418 & 0.2329 & 27.56 & 0.8399 & 0.6059  \\
    & & & & & & & & & & 13.39 & 0.8464 & 0.2217 & 28.61 & 0.8440 & 0.6362 \\ 
    \bottomrule[1.5pt]
    \end{tabular}
    }
    \vspace{-3pt}
    \caption{
    \textbf{Ablations on SceneGen Variants.} 
    We progressively remove global geometric features~($\mathcal{F}_{\mathrm{global}}^{\mathrm{geo}}$), global visual features~($\mathcal{F}_{\mathrm{global}}^{V}$), mask visual features~($\mathcal{F}_i^{\mathrm{mask}}$), and substitute the scene-level self-attention~($\mathcal{A}_{\mathrm{SS}}$) to validate each component's contribution to SceneGen.
    }
    \label{tab:ablation_study}
    \vspace{-8pt}
\end{table*}

\subsection{Qualitative Results}
\label{subsec:qualitative_results}
\noindent \textbf{Comparisons with baselines.}
As illustrated in Figure~\ref{fig:qualitative_results}, we qualitatively compare SceneGen with existing baselines on both the 3D FUTURE~\cite{fu20203dfuture} test set and in-the-wild ScnaNet++~\cite{yeshwanth2023scannet++}, where they still struggle with 3D scene generation: 
PartCrafter lacks controllability over the generated targets and often mistakenly merges distinct assets, while both PartCrafter and DepR are limited to geometry generation and cannot render textures. 
More critically, all these methods exhibit difficulties in accurately understanding the spatial relationships among assets.
In contrast, our proposed SceneGen precisely predicts the spatial relationships among assets and synthesizes multiple 3D assets with accurate geometry and high-quality textures, without relying on any additional tools or optimizations.

\vspace{2pt}
\noindent \textbf{Extension to multi-image inputs.}
Benefiting from our architecture design, SceneGen can seamlessly handle multi-image inputs after being trained exclusively on single-image samples.
Given the lack of suitable datasets for quantitative evaluation, we qualitatively assess the impact of multi-image inputs by randomly sampling several scenes from ScanNet++~\cite{yeshwanth2023scannet++} and employing SAM2~\cite{ravi2024sam2} to obtain segmentation masks of corresponding objects.
As depicted in Figure~\ref{fig:extension_to_multi_image}, compared to single-image inputs, incorporating multi-view images leads to 3D assets with more complete geometry and finer texture details.
This illustrates that SceneGen can adaptively integrate complementary information from multiple views to produce higher-quality 3D scenes, further validating its practicality and scalability.
More qualitative results will be included in Sec~\ref{subsec:extension_to_multi_image} of the \textbf{Appendix}.

\subsection{Ablation Studies}
\label{subsec:ablation_studies}
To validate the efficacy of our modules, we conduct comprehensive evaluations on several variants of SceneGen, assessing both the geometric and visual quality of synthesized scenes.
Concretely, we investigate the impact of gradually removing global geometric features~($\mathcal{F}_{\mathrm{global}}^{\mathrm{geo}}$), global visual features~($\mathcal{F}_{\mathrm{global}}^{V}$), mask visual features~($\mathcal{F}_i^{\mathrm{mask}}$), as well as substituting the scene-level self-attention block~($\mathcal{A}_{\mathrm{SS}}$) with a simple asset-level self-attention block~($\mathcal{A}_{\mathrm{AS}}$).
As depicted in Table~\ref{tab:ablation_study}, we have the following observations:
(i) Removing any of the aforementioned components degrades the overall performance, confirming their necessity in SceneGen;
(ii) The geometric features primarily affect the structure of synthesized scenes, while the visual features further impact the visual quality;
and
(iii) The absence of scene-level self-attention blocks eliminates inter-asset interactions during generation, leading to notable performance declines across all metrics.
These results strongly demonstrate the necessity and effectiveness of our proposed feature extraction and aggregation modules for SceneGen.

\section{Conclusion}
\label{sec:conclusion}
In this paper, we present \textbf{SceneGen}, a novel framework that takes a single scene image and target object masks as input to simultaneously synthesize multiple 3D assets with structure, texture, and relative spatial positions in a single feedforward pass.
Specifically, we incorporate dedicated visual and geometric encoders to extract both asset-level and scene-level features, which are effectively fused with our proposed feature aggregation module.
Notably, through our meticulous design, SceneGen can naturally generalize to multi-image inputs and achieve even better generation fidelity.
Quantitative and qualitative evaluations demonstrate that SceneGen produces physically plausible and mutually consistent 3D assets, significantly outperforming previous methods in terms of generation quality and efficiency.
\section*{Acknowledgments}
\vspace{-0.1cm}
Weidi would like to acknowledge the funding from Scientific Research Innovation Capability Support Project for Young Faculty~(ZY-GXQNJSKYCXNLZCXM-I22).

\clearpage

{
    \small
    \bibliographystyle{ieeenat_fullname}
    \bibliography{main}
}


\onecolumn
{
    \centering
    \Large
    \textbf{SceneGen: Single-Image 3D Scene Generation in One Feedforward Pass} \\
    \vspace{0.5em} Appendix \\
    \vspace{1.0em}
}

\appendix
{
  \hypersetup{linkcolor=black}
  \tableofcontents
}

\clearpage

\section{Preliminaries on 3D Foundation Models}
\label{sec:preliminary}
Given the inherent challenges of directly generating a 3D scene with multiple 3D assets from a single image, SceneGen aims to fully leverage the visual and geometric priors embedded in state-of-the-art 3D foundation models.
Therefore, we build our model based on TRELLIS~\cite{trellis}, and adopt DINOv2~\cite{dinov2} and VGGT~\cite{wang2025vggt} as our visual and geometric encoders, respectively.
In the following, we provide a detailed introduction to TRELLIS and VGGT to better illustrate their roles.

\vspace{3pt}
\noindent \textbf{TRELLIS.}
For a 3D asset~($\mathcal{O}$), TRELLIS encodes its geometry and appearance into a unified representation~($\boldsymbol{z}$), denoted as: $\boldsymbol{z} = \{(\boldsymbol{z}_i,\boldsymbol{p}_i)\}_{i=1}^{L}$.
Here, $\boldsymbol{p}_i\in \{0, 1,\ldots, D-1\} ^3$ denotes the positional index of an active voxel in the 3D grid intersecting the surface of $\mathcal{O}$, and $\boldsymbol{z}_i\in\mathbb{R}^C$ represents the local latent feature attached to the corresponding voxel, with $D$ and $L$ representing the 3D grid resolution and the total number of active voxels, respectively.

The generation process adopts two cascaded rectified flow models: 
the \textbf{sparse structure generator}~($\boldsymbol{\mathcal{G}}_{\mathrm{S}}$) synthesizes the sparse voxel structure $\{\boldsymbol{p}_i\}_{i=1}^{L}$, encoding geometric priors by predicting its low-resolution feature gird~($\boldsymbol{S}$); 
while the \textbf{structured latents generator}~($\boldsymbol{\mathcal{G}}_{\mathrm{L}}$) generates texture and appearance features $\{\boldsymbol{z}_i\}_{i=1}^{L}$ conditioned on $\{\boldsymbol{p}_i\}$. 
Both models are optimized via the conditional flow matching~(CFM)~\cite{lipman2023flowmatching} objective, which establishes straight probability paths between distributions through linear interpolation: 
$\boldsymbol{x}(t) = (1-t)\boldsymbol{x}_0 + t\boldsymbol{\epsilon}$, where $\boldsymbol{x}_0$ denotes data samples, $\boldsymbol{\epsilon}\sim \mathcal{N}(0, \mathbf{I})$, and $t \in [0,1]$.
The velocity field~($\boldsymbol{v}(\boldsymbol{x},t) = \nabla_t\boldsymbol{x}$) governs the reverse process, with the CFM objective formulated as:
\begin{align*}
    \mathcal{L}_{\mathrm{cfm}}(\theta)=\mathbb{E}_{t,\boldsymbol{x}_0,\boldsymbol{\epsilon}}\|\boldsymbol{v}_\theta(\boldsymbol{x}, t)-(\boldsymbol{\epsilon}-\boldsymbol{x}_0)\|^2_2
\end{align*}

Notably, the sparse structured generator~($\boldsymbol{\mathcal{G}}_{\mathrm{S}}$) learns rich geometric priors from large-scale 3D data, effectively capturing both object geometries and spatial relationships, thus delivering essential asset-level understanding capabilities.
Our SceneGen is compatible with both the sparse structured generator and structured latents generator, thus can sequentially employ them to decode synthesized latent features into the geometry and texture of 3D assets.

\vspace{3pt}
\noindent \textbf{VGGT.}
Trained on large-scale 3D annotated data, VGGT can extract 3D scene features through a purely feedforward network without explicit 3D inductive biases.
For single or multi-view RGB inputs~($\{\boldsymbol{I}_i\}_{i=1}^{s}$), its aggregator derives scene geometric features~($\{\mathcal{F}_i^{\mathrm{geo}}\}_{i=1}^{s}$), represented as:
\begin{align*}
    \{\mathcal{F}_i^{\mathrm{geo}}\}_{i=1}^{s} = \{[\mathcal{F}_{G}^{\mathrm{geo}}, \mathcal{F}_{I}^{\mathrm{geo}}]\}_{i=1}^{s} = \mathrm{VGGT}(\{\boldsymbol{I}_i\}_{i=1}^{s})
\end{align*}
Here, $\mathcal{F}_{G}^{\mathrm{geo}}$ and $\mathcal{F}_{I}^{\mathrm{geo}}$ denote features extracted by {\em global self-attention} and {\em local self-attention} layers, respectively.
These features are efficiently decoded by lightweight DPT layers~\cite{DPT} into depth maps, point maps, and tracks, validating their rich scene geometric representation capacity.

By integrating these complementary strengths, our \textbf{SceneGen} effectively captures both local asset-level and global scene-level features from the input image, achieving robust performance on the challenging 3D scene generation task.

\section{More Details about Training Data}
\label{sec:more_data_details}
\begin{figure}[t]
	\centering
	\includegraphics[width=\linewidth]{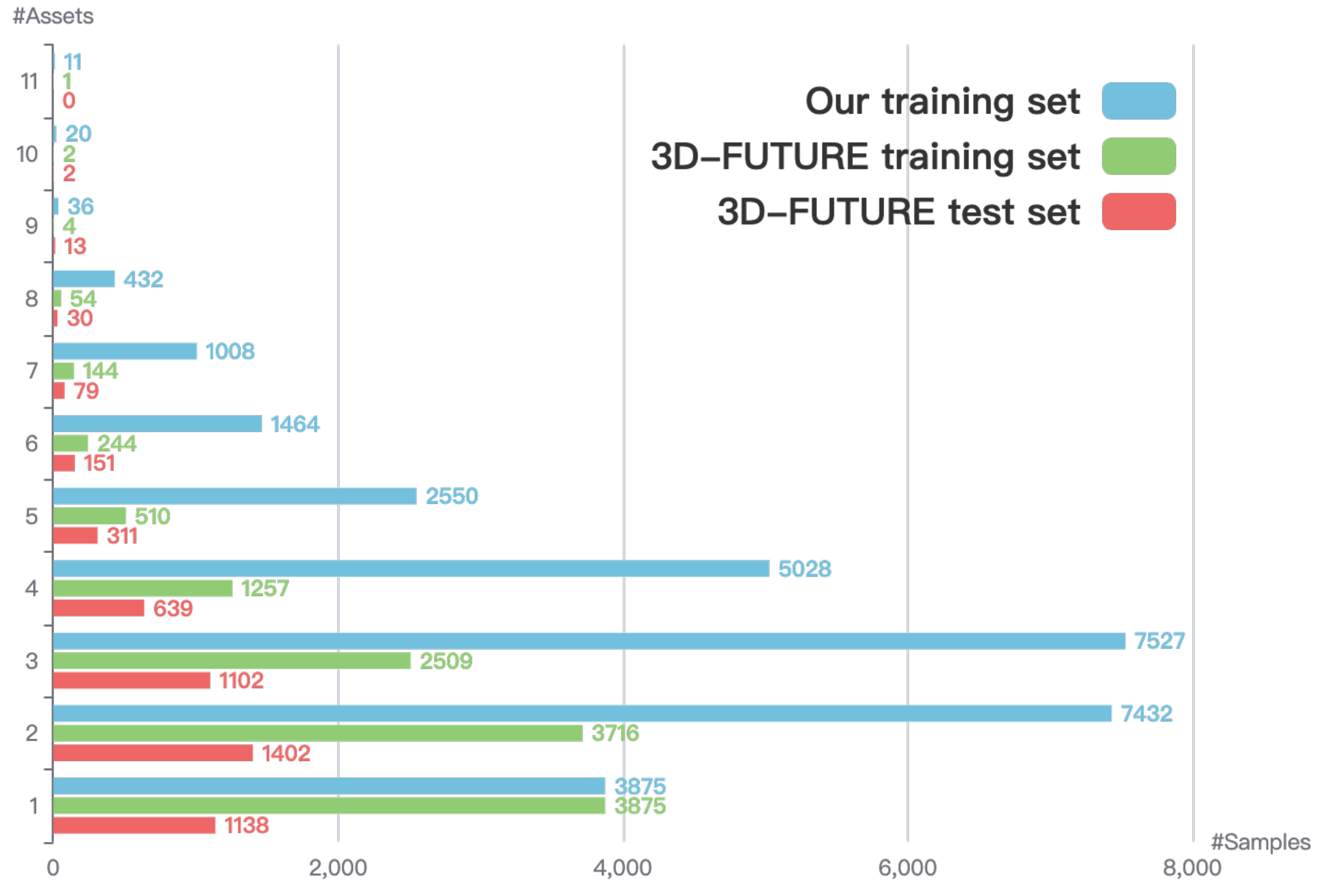}
    \vspace{-15pt}
    \caption{
        \textbf{Distribution of Asset Counts in our Training Data and Original 3D-FUTURE.}
    }
    \label{fig:dataset}
    \vspace{-9pt}
\end{figure}

We train SceneGen on the 3D-FUTURE~\cite{fu20203dfuture} dataset, leveraging its rich textures and diverse lighting conditions to simulate real-world environments and thereby enhance the model's generalization ability.
Additionally, to ensure the model robustly learns the relative spatial relationships among multiple assets, we further scale up training data through data augmentation.
Specifically, for a scene with $N$ objects, we iteratively select each asset as the query asset and randomly shuffle the remaining ones during training.
Considering GPU memory constraints, we set the maximum number of assets per scene to $N' = 7$ on a single A100 GPU.
For samples containing more than $N'$ assets, we randomly select a subset of $N'$ assets for training. 
Furthermore, following TRELLIS~\cite{trellis}, we apply its aesthetic score filtering criterion to exclude assets with aesthetic scores below 4.5, thereby ensuring high data quality.
The distribution of asset counts across training scenes is illustrated in Figure~\ref{fig:dataset}.

\section{More Implementation Details}
\label{sec:more_implementation_details}
This section provides a holistic explanation of implementation details discussed in the paper.
Concretely, Sec.~\ref{subsec:extension_to_multi_image} describes the specific strategies applied to extend SceneGen to multi-image inputs; and Sec.~\ref{subsec:evaluation_protocols} elaborates on our evaluation protocols.

\subsection{Extension to Multi-image Inputs}
\label{subsec:extension_to_multi_image}
While SceneGen is primarily designed for 3D scene generation based on a single scene image and trained exclusively on single-view images, it can be seamlessly adapted to multi-view inputs during inference with no need for additional training or fine-tuning. 
Concretely, during inference, our model can take images of the same scene from multiple viewpoints, along with their corresponding objects and instance masks, as input.
Within our SceneGen, the geometric encoder~($\Phi_{G}$)~(an off-the-shelf VGGT~\cite{wang2025vggt} aggregator) integrates geometric information across different viewpoints to produce better geometric representations for each perspective, thereby enabling SceneGen to synthesize more accurate geometric structures. 
Finally, we predict the relative positions among different assets from each viewpoint and use the mean of these predictions across all views as the final spatial position output. 
It is important to note that, to ensure correctness throughout the inference process, the input order of assets and their segmentation masks must remain consistent across all viewpoints.

Given the current lack of training and quantitative evaluation data for multi-view 3D scene generation, this work presents qualitative results on scenes sampled from ScanNet++~\cite{yeshwanth2023scannet++} to demonstrate the scalability of SceneGen, and leaves the construction of suitable multi-view datasets and evaluation methods for future work.

\begin{table}[hbpt]
    \centering
    \small
    \setlength{\tabcolsep}{3.3pt} 
    \renewcommand{\arraystretch}{1.1}
    \centering
    \begin{tabular}{c|c|ccccc}
        \toprule
         \textbf{Method} & \textbf{Alignment} & \textbf{CD-S}$\downarrow$ & \textbf{CD-S 1}$\downarrow$ & \textbf{CD-S 2}$\downarrow$ & \textbf{F-Score-S}$\uparrow$ & \textbf{IoU-B}$\uparrow$ \\
        \midrule
        \multirow{2}{*}{MIDI~\cite{huang2025midi}} & ICP & 0.1697 & 0.0653 & 0.1044 & 41.64 & 0.1232 \\
        & FilterReg & 0.0501 &  0.0278 & 0.0223 & 68.74 & 0.2493 \\
        \midrule
        \multirow{2}{*}{\centering \makecell{\textbf{SceneGen~(Ours)}}} & ICP & 0.0310 & 0.0121 & 0.0189 & 83.74 & 0.5103 \\
        & FilterReg & \textbf{0.0118} & \textbf{0.0052} & \textbf{0.0066} & \textbf{90.60} & \textbf{0.5818} \\
        \bottomrule
    \end{tabular}
    \vspace{-3pt}
    \caption{
    \textbf{Geometric Metrics Comparisons on Different Point Cloud Alignment Methods.}
    }
    \vspace{-9pt}
    \label{tab:alignment_difference}
\end{table}

\begin{figure}[t]
	\centering
	\includegraphics[width=\linewidth]{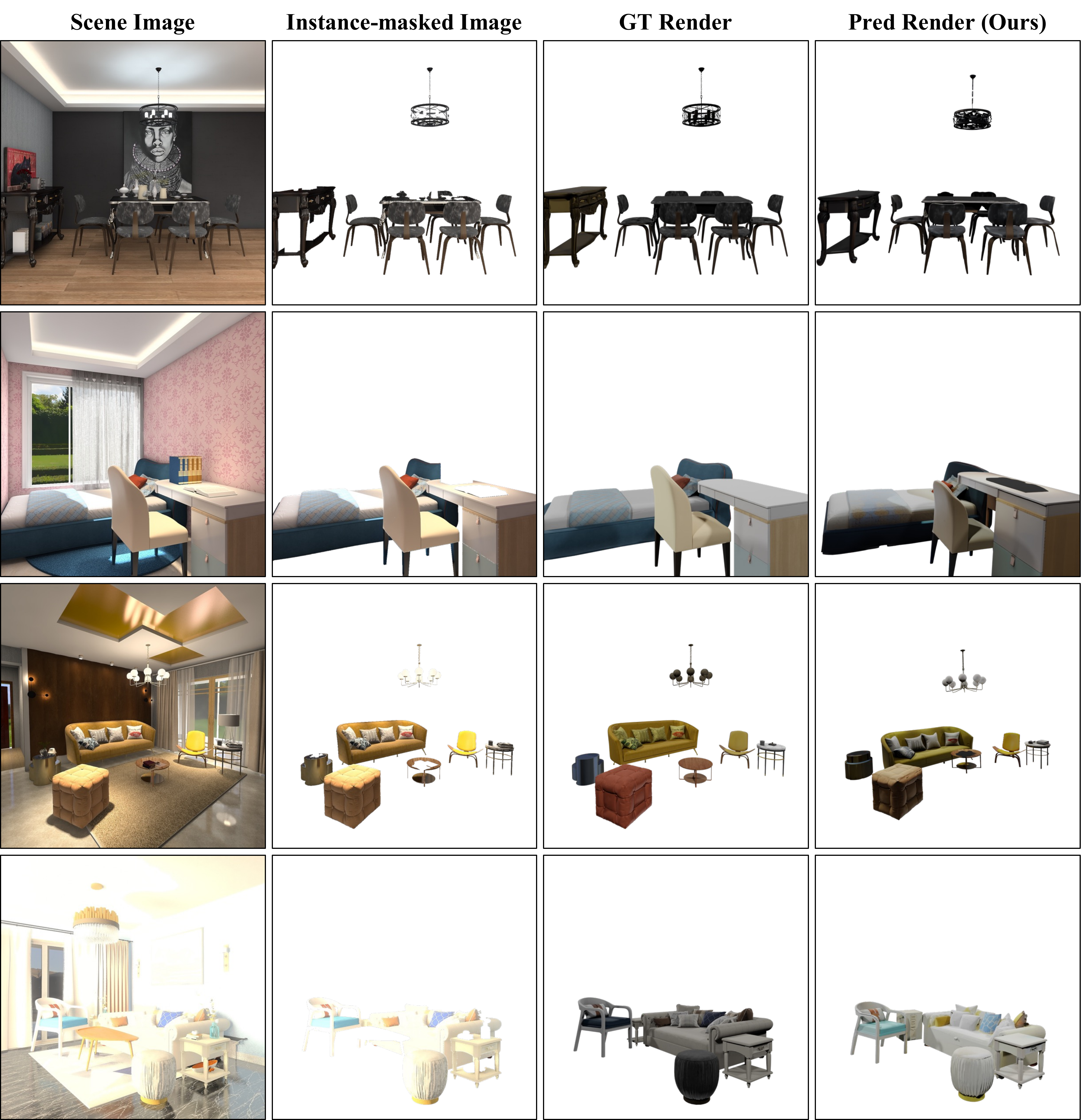}
    \vspace{-9pt}
    \caption{
    \textbf{Examples of Visual Metrics Evaluation Protocols.}
    Here, we present two complementary types of ground truth: instance-masked images may introduce slight differences due to potential occlusions, while GT-render images lack scene-level illumination.
    }
    \label{fig:visual_metric}
    \vspace{-9pt}
\end{figure}

\subsection{Evaluation Protocols}
\label{subsec:evaluation_protocols}
\noindent \textbf{Geometric metrics.}
Following previous work~\cite{huang2025midi}, we conduct geometry evaluation in normalized 3D space~(also referred to as canonical space, {\em i.e.}, $x, y, z \in [-1, 1]$), where the ground truth and the synthesized query asset are first rigidly aligned using point cloud registration algorithms.
Unlike MIDI~\cite{huang2025midi}, which relies on the traditional Iterative Closest Point~(ICP~\cite{ICP}) method prone to suboptimal alignment results, we employ FilterReg~\cite{gao2019filterreg}, a faster and more robust point cloud alignment approach.
As presented in Table~\ref{tab:alignment_difference}, both MIDI and SceneGen achieve better overall performance when aligned via FilterReg, demonstrating the reliability of this alignment method compared to traditional ICP.
Moreover, under both alignment strategies, SceneGen consistently outperforms MIDI, indicating that explicitly predicting the spatial positions among assets enables SceneGen to more accurately model the relationships among distinct 3D assets within the scene.

\vspace{3pt}
\noindent \textbf{Visual metrics.} 
Beyond the commonly used geometric evaluations described above, we also consider several visual metrics to assess the visual quality of generated scenes. 
Concretely, after aligning the synthesized point clouds with the ground truth scenes, we use {\em Blender} to render them with the identical camera parameters.
The rendered images are then compared with two types of ground truth to compute perceptual metrics that reflect the visual quality of synthesized scenes.
As illustrated in Figure~\ref{fig:visual_metric}, these include:
(i) instance-masked scene images, which are extracted using the corresponding object masks, where the occlusion relationships between assets introduce differences relative to predicted renderings;
and 
(ii) GT-Render images, which are rendered from the ground truth assets at the same viewpoint using {\em Blender}, but lack scene-level illumination and complete textures, resulting in textural discrepancies compared to predicted scenes.
Thus, by computing visual metrics against both types of ground truth, we provide a complementary evaluation of the visual quality of synthesized scenes.

\vspace{3pt}
\noindent \textbf{Efficiency.}
To ensure a fair comparison across all methods, we report the average inference time over 500 trials of synthesizing scenes with a single asset on a single A100 GPU.
Notably, our proposed SceneGen can directly generate 3D scenes containing 4 assets in a single feedforward pass within 2 minutes on the same hardware, eliminating the need for time-consuming sequential generation of individual 3D assets.

\begin{figure*}[t]
  \centering
  \includegraphics[width=\textwidth]{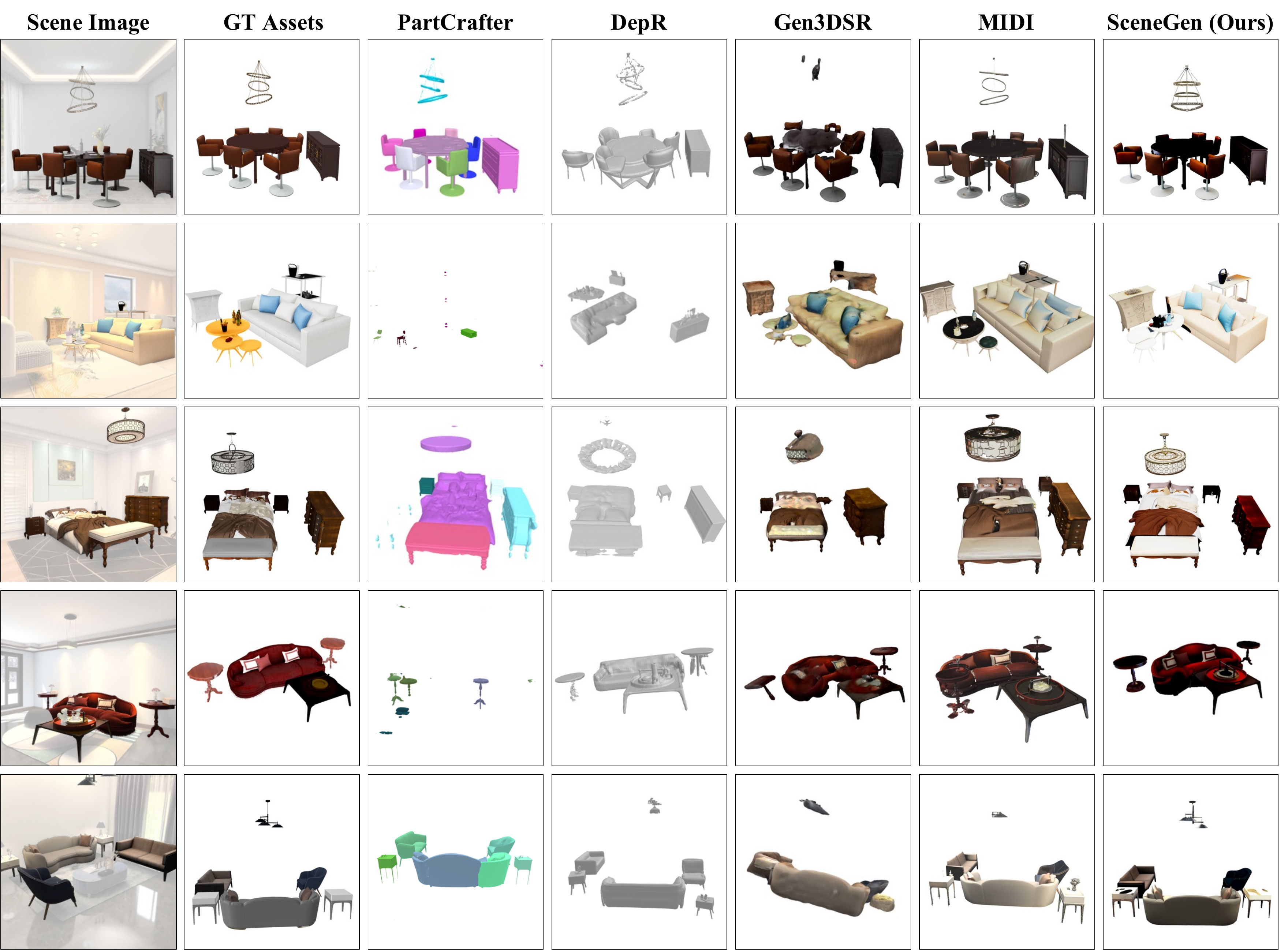} \\
  \caption{
    \textbf{More Qualitative Comparisons on the 3D FUTURE Test Set.} 
  }
    \vspace{-9pt}
 \label{fig:qualitative_results_sup}
\end{figure*}

\section{More Visualizations}
\label{sec:more_visualizations}
This section presents additional qualitative results on the 3D-FUTURE~\cite{fu20203dfuture} test set, offering a detailed comparison between our SceneGen and representative baselines.
As depicted in Figure~\ref{fig:qualitative_results_sup}, we have the following observations:
(i) PartCrafter~\cite{lin2025partcrafter} frequently suffers from missing or mixed-up assets due to its inability to control generation via object masks, despite already taking segmented objects and asset counts as input;
(ii) Both PartCrafter and DepR~\cite{zhao2025deprdepthguidedsingleview} can only generate scene geometry without rendering texture details;
and 
(iii) All baseline methods~(PartCrafter~\cite{lin2025partcrafter}, DepR~\cite{zhao2025deprdepthguidedsingleview}, Gen3DSR~\cite{dogaru2025gen3dsr}, and MIDI~\cite{huang2025midi}) share the common limitation of incorrect spatial relationships among synthesized assets.
In contrast, our SceneGen fully integrates visual and geometric features within the scene to enable mutual influence among multiple assets during generation, producing 3D scenes with physically plausible geometry and high-quality texture details.

\section{Limitations \& Future Works}
\subsection{Limitations}
While our SceneGen demonstrates superior performance in 3D scene generation, it is not without its limitations.

\vspace{2pt}
\noindent \textbf{Limited to Indoor Generation.}
While SceneGen demonstrates better texture generation and generalization capabilities compared to previous methods that rely on canonical representations, the narrow training data distribution limits its ability to generalize to non-indoor scenes, restricting its generalization to a broader range of environments.

\vspace{2pt}
\noindent \textbf{Asset Collisions and Overlaps.}
Although SceneGen can generate multiple 3D assets and relative spatial positions in a single feedforward pass, without relying on complex post-processing, it does not always handle contact relationships among objects, occasionally leading to asset overlaps or geometric inconsistencies. 
This is mainly because our single-stage framework does not explicitly enforce strict spatial or physical constraints among objects.

\vspace{2pt}
\noindent \textbf{Reliance on Segmentation Masks.}
SceneGen inherently requires segmentation masks of the target objects as input. 
In our current framework, we leverage either ground truth masks or masks pre-extracted by an off-the-shelf SAM 2~\cite{ravi2024sam2}. 
This reliance limits the flexibility of applying SceneGen directly to in-the-wild data to some extent and may potentially result in a lack of robustness against low-quality segmentation masks.

\subsection{Future Works}
To address the aforementioned limitations of SceneGen, we propose several directions for future improvement:
(i) Constructing larger-scale 3D scene generation datasets that cover more diverse indoor and outdoor scenarios, to address biases in training data distribution and improve the generalization ability of models;
(ii) Building suitable multi-view scene generation datasets to expand the application scope and practical potential of existing models;
(iii) Incorporating explicit physical priors or constraints to facilitate the model to better learn complex interactions among objects;
and
(iv) Introducing an additional object segmentation module into the current framework or natively integrating the capability to segment assets from scenes.

\end{document}